\documentclass{article}

\usepackage{amssymb, amsmath, amsopn}
\usepackage{times,mathptmx}
\usepackage{mathptmx}
\usepackage{helvet}
\usepackage{courier}
\usepackage{makeidx}
\usepackage{multicol}
\usepackage{footmisc}
\usepackage{amssymb, amsmath}
\usepackage{graphicx}
\usepackage{caption}
\usepackage{subfig}
\usepackage{xspace}
\usepackage{color}
\usepackage{array}
\usepackage{verbatim}
\usepackage{imakeidx, epsfig,url}
\usepackage{xspace}
\usepackage{balance}
\usepackage{graphicx}

\usepackage{mathptmx}
\usepackage{helvet}
\usepackage{courier}
\usepackage{makeidx}
\usepackage{multicol}
\usepackage{footmisc}
\usepackage[justification=centering]{caption}
\usepackage{amssymb, amsmath}
\usepackage{graphicx}
\usepackage{caption}
\usepackage{subfig}
\usepackage{xspace}
\usepackage{color}
\usepackage{array}
\newcolumntype{P}[1]{>{\centering\arraybackslash}p{#1}}
\usepackage{verbatim}
\usepackage{imakeidx, epsfig,url}
\usepackage{makeidx}   
\usepackage[lined,boxed,linesnumbered]{algorithm2e} 
\begin{document}

\title{A Feature-Based Prediction Model of Algorithm Selection for Constrained Continuous Optimisation}
\author{
	Shayan Poursoltan\\
	Optimisation and Logistics\\
	School of Computer Science\\
	The University of Adelaide\\
	Adelaide, Australia
	\and 
	Frank Neumann\\
	Optimisation and Logistics\\
	School of Computer Science\\
	The University of Adelaide\\
	Adelaide, Australia
}

\date{}
\maketitle
\begin{abstract}
	With this paper, we contribute to the growing research area of feature-based analysis of bio-inspired computing. In this research area, problem instances are classified according to different features of the underlying problem in terms of their difficulty of being solved by a particular algorithm. We investigate the impact of different sets of evolved instances for building prediction models in the area of algorithm selection. Building on the work of Poursoltan and Neumann~\cite{Poursoltaniconip1,Poursoltaniconip2}, we consider how evolved instances can be used to predict the best performing algorithm for constrained continuous optimisation from a set of  bio-inspired computing methods, namely high performing variants of differential evolution, particle swarm optimization, and evolution strategies. Our experimental results show that instances evolved with a multi-objective approach in combination with random instances of the underlying problem allow to build a model that accurately predicts the best performing algorithm for a wide range of problem instances.
\end{abstract}

\section{Introduction}
Throughout the history of heuristic optimisation, various methods have been proposed to solve constrained optimisation problems (COPs), specially non-linear ones. The main idea behind these algorithms is to tackle the constraints. Important approaches in this area are differential evolution (DE), particle swarm optimisation (PSO) and evolutionary strategies (ES). To handle the constraints, there have been many techniques applied to these algorithms such as penalty functions, special operators (separating the constraint and objective function treatment) and decoder based methods. We refer the reader to \cite{mezura2011constraint} for a survey of constraint handling techniques in evolutionary computation. Given a range of different algorithms for constrained continuous optimisation, we consider algorithm selection problem (ASP)~\cite{rice1975algorithm} which consists of selection the best performing algorithm from a suite of algorithms for a given problem instance. In most circumstances, it is difficult to answer the following question: "Can we estimate the likelihood that algorithm $A$ will be successful on a given constrained optimisation problem $P$?". Recent works in the field show that it is possible to select the algorithm most likely to be best suited for a given problem \cite{munoz2012meta,smith2008towards,bischl2012algorithm}. Based on these studies, it is possible to find the links between problem characteristics and algorithm performance. The key to these investigations is problem features which can be used to predict the most suited algorithm from a set of algorithms.

It is widely assumed that constraints play a vital role in COP's difficulty.  Therefore, in this study we use the meta-learning framework outlined in \cite{rice1975algorithm,smith2009knowledge} to build a prediction model for a given COP. Our model predicts the best algorithm type (DE, ES and PSO) for a given COP based on their constraint features. The model inputs include the features of  constraints in a given problem. It is shown in \cite{Poursoltaniconip1,poursoltan2014feature} that by using an evolving approach, it is possible to generate problem instances covering a wide range of problem/algorithm difficulty. Such instances can be used to extract and analyse the features that make a problem hard or easy to solve for a given algorithm. For a detailed discussion on these constraints (linear, quadratic and their combination) we refer the reader to~\cite{Poursoltaniconip1}. 

To build a reliable prediction model, we need to train it with variety of problem instances that are hard or easy for algorithm(s). Based on the investigations in~\cite{Poursoltaniconip2}, a multi-objective evolutionary algorithm can be used to generate constrained problem instances that are hard/easy for one algorithm but still easy/hard for the others. The authors show which features of the constraints make the problems hard for certain algorithm but still easy for the others. Hence, we use the same approach to generate problem instances to use in our model training phase. This can improve the accuracy of prediction model since the training instances are used to show the strengths and weaknesses of various algorithm types over constraint features. To illustrate the model's efficiency on constraints, we examine our model with generated testing problems such as hard/easy for one but easy/hard for the others and more general random instances. To show the model prediction ability over constraints (linear, quadratic and their combination), we also experiment given problems with various objective functions.

The remainder of this paper is organised as follows. In section 2, we discuss constrained continuous optimisation problems. Later, we introduce the evolutionary algorithms that are suggested by our prediction model. Moreover, the background materials related to multi-objective evolver, algorithm selection problem and meta-learning prediction model are discussed in detail. Section 3 describes and compares all models trained with different subsets of instances from multi-objective evolver population set. By choosing the best training data preference in Section 3, the experimental analysis on various benchmark problems is described in Section 4. We then conclude with some remarks in Section 5.

\section{Preliminaries}
\subsection{Constrained Continuous Optimisation Problems}
A constrained optimisation problem (COP) in a continuous space is formulated as follows:
\begin{equation}
	\begin{split}
		\quad \text{Find} &\quad x \in S  \subseteq R^{D}\\
		&\quad    f(x) = min(f(y);y \in S), \\
		\text{subject to}&\quad	g_i(x)  \leq 0 \quad  \forall   i \in \{1,\ldots,q\}\\
		&	\quad	h_j(x) = 0 \quad \forall j \in \{q + 1,\ldots, p\} 
		\label{eq:f}
	\end{split}
\end{equation}
In this formulation, $f$, $g_{i}$ and $h_{j}$ are real-valued functions on the search space $S$, $q$ is the number of inequalities and $p-q$ is the number of equalities. The search space $S$ is defined as a $D$ dimensional rectangle in $R^{D}$. These equality and inequality constraints could be linear or nonlinear. The set of all feasible points $F \subseteq S$ which satisfy all equality and inequality constraints is formulated as:
\begin{equation}
	\
	l_i  \leq x_i  \leq u_i,\quad  \quad     1 \leq i  \leq D
	\label{equality}
\end{equation}
where $l_i$  and $u_i$ denote lower and upper bounds for the $i$th variable respectively. Usually, to simplify COP, the equalities are replaced by the following inequalities \cite{takahama2010efficient} as follows:

\begin{equation}
	|h_j(x)|  \leq \epsilon \quad \text{for} \enspace j=q+1 \enspace \text{to} \enspace  p
	\label{transform}
\end{equation}
where $\epsilon$ is a small positive value. In all experiments in this paper,
the value of $\epsilon$ is considered as 1E-4, the same as it was in \cite{mallipeddi2010problem}. 

\subsection{Algorithms}
In this section we  discuss the basic ideas about algorithms for constrained optimisation problems such as differential evolution, evolutionary strategies and particle swarm optimisation. 

The $\epsilon$-constrained differential evolution with an archive and gradient-based mutation ($\epsilon$DEag) is the winner of 2010 CEC competition for continuous constrained optimisation problems \cite{mallipeddi2010problem}. This algorithm uses $\epsilon$-constrained method technique to transform  algorithms for unconstrained problems to constrained ones. Also, possible solutions are ordered by $\epsilon$-level comparison. This means, the lexicographic order is used in which constraint violation ($\phi(x)$) has more priority and proceeds the function value ($f(x)$). A detailed description of this algorithm can be found in \cite{takahama2010constrained}.

For evolutionary strategy algorithms, $(1+1)$ CMA-ES for constrained optimisation \cite{arnold20121+} is included in our experiment. This algorithm is a variant of $(1+1)$-ES which adapts the covariance matrix of its offspring distribution in addition to its global step size. The $(1+1)$ CMA-ES for constrained optimisation obtains approximations to the normal vectors directions in the vicinity of the current solution locations by applying low-pass filtering steps that violates the constraints and reducing the variance of the offspring distribution in these directions. Adopting this method makes $(1+1)$ CMA-ES as one of the most efficient algorithms for constrained optimisation problems. We refer the reader to \cite{arnold20121+} for detailed description and implementation.

The next algorithm that is used in our investigation from particle swarm optimisation algorithms is hybrid multi-swarm particle swarm optimisation (HMPSO). This algorithm divides the current swarms into sub-swarms and search the solution between them in parallel. All particles in each sub-swarms locate their fittest local particle which attracts the particles to fitter positions. Also, having multiple sub-swarms near different optima increase the diversity of the algorithm. A detailed description and implementation of HMPSO can be found in \cite{wang2009hybrid}.

\subsection{Multi-objective Investigations}
\label{Multi_investi}
In order to extract information about the strengths and weaknesses of certain algorithms on constrained optimisation problems, we need problem instances with different kinds of difficulties for the considered algorithms. The reason behind this idea is that using instances that are randomly generated are not efficient to cover the full spectrum of difficulty analysis. To do this, we evolve instances to find the ones that are hard/easy for one algorithm and easy/hard for the others. Analysing the features of these instances helps us extracting knowledge regarding the strengths and weaknesses of the considered algorithms and give reasons of why an algorithm performs well on one problem while the others have difficulties. Insights from this analysis can be used to develop more efficient prediction model for automated algorithm selection.

As mentioned above, we generate problem instances of different difficulties. To do this, a multi-objective DE algorithm (evolver) \cite{robivc2005demo} is used to evolve constraints that make problems hard for one algorithm type and easy for the others in the algorithm suite following the approach in~\cite{Poursoltaniconip1,Poursoltaniconip2}. The feature-based comparison of various algorithm types has been carried out in these papers. The authors show the constraint (linear and/or quadratic and their combination) features that are more contributing to problem difficulty for certain algorithms. We refer the readers to \cite{Poursoltaniconip1,Poursoltaniconip2} for a detailed description and implementation.

\subsection{Algorithm Selection Problem}
\begin{figure}

	\centering		
	\includegraphics[scale=0.7]{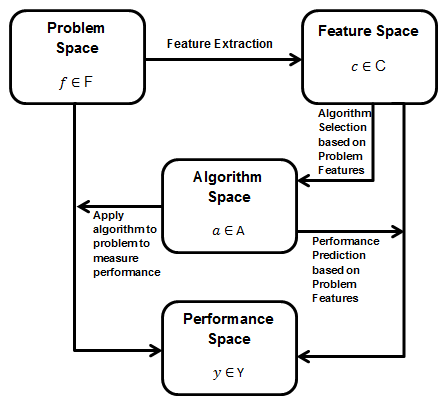}
	\caption{
		The framework of general problem of algorithm selection and performance prediction using problem features based on \cite{rice1975algorithm}.}
	\label{fig:algorithmselection}
\end{figure}

There are many algorithms that are proposed for constrained continuous optimisation problems. These algorithms are categorised as different types such as differential evolution (DE), evolutionary strategy (ES) and particle swarm optimisation (PSO). So, as a direct consequence of this, it is difficult to understand which algorithms or types of algorithms are more efficient to solve given COPs. To determine the best algorithm to solve a problem is referred to "Algorithm Selection Problem" term in \cite{rice1975algorithm} by Rice. In his work, Rice proposed a model with four main characteristics: a set of problem instances $F$, a set of algorithms $A$, measures for the cost of performing algorithms on particular problem ($Y$) and set of characteristics of problem instances ($C$). The illustration for Rice general algorithm selection framework is shown in Figure \ref{fig:algorithmselection} which predicts the performance $y(a(f))$ of a given algorithm $a$ on a problem $f$ by extracted features $c$. If a solution is found, it is possible to extract features from a given problem and select the most appropriate algorithm or predict the performance of the algorithm based on these features. This framework has been extended by \cite{smith2009knowledge,hutter2006performance,leyton2009empirical} in a variety of computational problem domains using meta-learning framework. So, if the values are features of problems with algorithm performance measure are known beforehand, then it is possible to use a learning strategy to predict the algorithm performance based on the problem features. 

\subsection{Prediction Model}
\begin{figure}

	\centering		
	\includegraphics[scale=0.6]{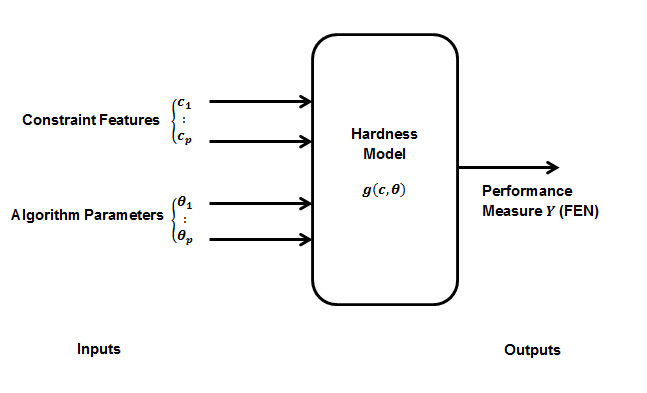}
	\caption{
		Meta-learning prediction model for a constrained continuous optimisation problem}
	\label{fig:prediction}
\end{figure}

Our prediction model is based on the \cite{munoz2012meta}. Note that this model is used for unconstrained continuous optimisation problems using landscapes features. Inputs to the model are independent problem feature variables ($C$) and algorithm parameters and output is the performance measure as an dependant variable which is required function evaluation number (FEN) of the suggested algorithm. This model can be used to predict the algorithm behaviour on a given problem. To achieve this we use popular basic technique for model building. A high-level overview of the regression model is shown in Figure \ref{fig:prediction}.

As discussed above, there have been many attempts to train these prediction models with random generated or benchmark problem instances which could not fully include all problem instances with difficulty variations. To improve this, we cover the full spectrum of difficulty by evolving two sets of instances with extreme problem difficulties. These extreme difficulty instances are the ones which are hard for one algorithm and easy for the others or easy for one and still hard for the rest.

To build our regression model, we implement a multi-layered feed-forward neural network with 2 hidden layers and 10 neurons in each layer as the regression model. For training the model, we use a Levenbeg-Marqurd back-propagation algorithm \cite{marquardt1963algorithm} package using Matlab R2014b. To train this model, we use evolved instances that are generated from multi-objective evolver in \cite{Poursoltaniconip2}. The prediction model inputs are given COPs constraint features and algorithms parameter values (the parameters for the experimented algorithms are identical to \cite{takahama2010constrained,wang2009hybrid,arnold20121+}.)

\section{Prediction Model based on Evolved Instances} 
As mentioned earlier, our goal is to propose a reliable prediction model using constraint features. This reliability can be improved by choosing proper set of learning data. The accuracy of this prediction model depends on many factors such as the relevance of constraint features, the diversity of instances used to train the model and its training method. Therefore, to improve this, we train our prediction model obtained from multi-objective evolver described in \cite{Poursoltaniconip2}. These instances are hard for one algorithm but still easy for the other (or easy for one and hard for the other algorithms). Analysing these instances shows the effectiveness of constraint features that make a problem hard or easy for certain algorithms. This set up improves the accuracy of prediction which is evaluated on the capability to provide realistic ranking of different algorithm types performances. This can be done by comparing the required function evaluation numbers (FEN) needed by various algorithms to solve a given COP.

The prediction model uses constraint features (first input) to predict the best algorithm type for a given constrained optimisation problem. These constraint features are constraint coefficients relationships such as standard deviation, angle between constraint hyperplanes, feasibility ratio in vicinity of optimum and number of constraints. The details of these features are discussed in \cite{Poursoltaniconip1}. Also, for the second input, since selecting various algorithm parameters has different impact on algorithm ability to solve a given problem, we conduct an experimentally driven meta-learning approach which has been proposed by Smith-Miles in \cite{smith2008cross}. For our model, we use the parameters for DE, CMA-ES and HMPSO suggested in \cite{takahama2010constrained,wang2009hybrid,arnold20121+}.

Our goal of building a prediction model is to identify the best algorithm type for a given problem. Therefore, the model output predicts the most suited algorithm and required FEN to solve a given constrained problem. The suggested FENs denote numbers of function evaluation which are needed by different algorithms to solve a given COP.

In the following we train our prediction model with variety of instance subsets generated by the multi-objective evolver. We choose different combination of subsets of instances to maximise our prediction model accuracy upon a given constrained problem. These training phase instance subsets are selected from extreme points, Pareto front line, more random (general) solutions and combination of Pareto front and random points from multi-objective evolver solution population in \cite{Poursoltaniconip2}. We then compare the prediction accuracy for these prediction model with various training data preferences.

\subsection{Extreme Instances}
We first train our prediction model with an extreme instance subset which covers the extreme points of Pareto front in the multi-objective evolver population set. These extreme solutions are selected from evolved instances that are easiest/hardest for one and hardest/easiest for the other algorithms at the same time. The reason behind this selection is to assess the ability of our model to find the most suited algorithm for a given COP which is fairly hard for one or multiple algorithms. In other words, it is more beneficial to choose a best algorithm for a given COP in which it cannot be solved easily by certain other algorithms.

To determine the actual accuracy of our extreme point prediction model (EP-PM), we select 1500 extreme instances that are hard/easy for one and easy/hard for the other algorithms for its training phase. To analyse and test the quality of this prediction model, we use two sets of testing problem instances that we already know their best algorithm and required FEN. The first one is the set of problem instances that are hard/easy for one algorithm and easy/hard for the others. This set can improve the accuracy of EP-PM for given problems that fall into extreme-like evolved problem instances. However, it is very likely that the real world given COP is similar to other evolved  instance subset types. Therefore, as a second set, we use random (general) testing problem instances to analyse the EP-PM with a potential real world given problem. We need to mention that we already know about their best algorithm and required FEN. 

Our result for EP-PM  is summarised in Table \ref{table:extonlylin} for instances that are hard/easy for one and easy/hard for the other algorithms. Also, the results for testing random instances are shown in this table. Moreover, DE hard (1 C) denotes testing Sphere problem instance with 1 linear constraint that is hard for DE algorithm but still easy for the others. The information about actual and predicted algorithm and required FEN to solve a problem instance is indicated. The model not only suggests the best algorithm with its required FEN but also predicts the FENs required to solve a given problem with other algorithms (not the best ones).

Based on the results, EP-PM performs acceptable on extreme-like testing instances. This performance is acceptable on predicted best algorithm type and required FEN. Also, the error rates for using other algorithms (not the best ones) are still acceptable. But, analysing the testing random instances, it is obvious the model is not capable of predicting algorithm and FEN for these more general form instances. This means, the proposed model (EP-PM) is not accurate enough for randomly generated real world instances and the difference between predicted and actual FEN is considerable. Also, the error rate of other algorithm choices (not the best algorithm) is still high.

To summarize, although the EP-PM model performs fairly accurate on instances that are grouped into extreme points evolved instances, still needs improvement to handle other subsets (such as random generated instances). The likelihood of given COP which is more similar to random evolved instances are higher. Thus, this motivates us to examine other subsets from evolving algorithm population instances for our training phase. This could be moving along the Pareto front line and choosing more instances from this category.

\subsection{Pareto Front Instances}
\label{Paretofront}
It is shown that in order to improve the accuracy of our prediction model we need to include or select different varieties of evolved instance subsets for its training step. The idea behind this choice is to obtain a model that can predict more general forms of given constrained problems. Of course it is vital to predict best algorithm for a given problem which is considerably hard/easy for one and still easy/hard for the other algorithms, but we also need to include more forms of generality to our prediction model. So, we need to move along Pareto front line in multi-objective evolver population set for our learning phase. This could increase the ability of our model to predict algorithms for more general given COP which is not similar to extreme point instances.  

Given a total number of 3000 instances from evolver Pareto front line, we train our Pareto front prediction model (PF-PM). This preference could increase PF-PM ability to predict more general forms of given COPs. To compare the quality of our prediction model and other models with different learning phases, we use same extreme and random generated testing instances used in previous section. Results shown in Table \ref{table:Paretoonly} indicate an improving accuracy for random testing instances used for previous model EP-PM (using extreme instances). Looking at the Table \ref{table:Paretoonly}, we see that moving towards Pareto front line in evolver population set for choosing learning instances increases the accuracy of predicted FEN for predicted algorithm. Also, predicting FEN using other algorithms (not the best one) represents that the PF-PM is more accurate than EP-PM for testing random instances. This improvement is acceptable in algorithm type prediction but we still need to improve the predicted required FEN.

As discussed in this section, to improve the accuracy of prediction model, we chose instances of evolver from its Pareto front line solution population. In other words, Pareto front prediction model (PF-PM) has some strengths and weaknesses. Although its error rate for predicting correct algorithms is improved, there is still considerable difference between the actual required FENs and predicted ones. As testing instances are selected mostly from more general instances (not close to extreme points), we need to experiment other training instances subsets. In order to address more general form, based on results so far, our next move is to choose more random instances from evolver population set for our training set. This could result in increasing the accuracy of our model for more general forms of given COPs.

\subsection{Random Instances}

The initial prediction models discussed earlier (EP-PM and PF-PM) has some limitations. In other words, they are not accurate enough to predict more general forms of given COPs, specially when they are similar to evolver population instances except extreme and Pareto front points. The results for PF-PM show an increase in accuracy of prediction for testing COPs which are not similar to extreme points, but the predicted required FEN still needs an improvement. Our goal is to design a prediction model with an ability to predict all possible given COPs. These COPs are within the range of extreme to random like instances. Based on previous results, it is shown that moving from extreme to Pareto front line instances increases the model accuracy (see Section \ref{Paretofront}). Hence, to decrease the error rate for required FEN for more general testing COPs we choose only random instances for testing phase. These random instances are selected from evolver population set.

We use 3000 random instances from multi-objective evovler population set for our random only prediction model (RO-PM). To assess the accuracy of our random only prediction model we use the same testing instances applied to EP-PM and PF-PM. Table \ref{table:comparingmethods} indicates the actual and predicted FEN and algorithm of hard/easy and random testing instances for RO-PM. As it is observed, the random only prediction model (RO-PM) fails to predict suitable algorithm for a given COP. This failure include both predicted algorithm and required FEN. Results show that moving through random instances in evolver population and choose only random instances increase the number of incorrect predictions. Also, comparing to previous models, RO-PM accuracy is decreased for testing instances similar to extreme points (easy/hard instances).

It is shown that the accuracy of resulting model with Pareto front instances (PF-PM) is improved by selecting different subsets (Pareto front line) than extreme points. This improvement is analysed by experimenting more general form of testing COPs. Therefore, this motivated us to experiment only random solutions in order to build more accurate prediction model for given COPs. By selecting only random instances to train the new model, it is observed the predicted algorithm and required FEN is not accurate as Pareto front (PF-PM) and extreme points (EP-PM) models. It can be translated as excluding instances from Pareto front line for training step decreases the accuracy of prediction model. Also, the RO-PM model failed to predict testing instances which are similar to evolver extreme points (easy/hard instances). So, other possibility is to use a combination of both Pareto front and random instances from evolver population for model training.

\subsection{Pareto Front with Random Instances} \label{Paretorandomsec}
In previous sections we experimented three types of prediction models (EP-PM, PF-PM and RO-PM). These models have different prediction accuracy upon choosing different varies of evolver subsets for their training phases. Our goal of building a prediction model is to minimise its error rate in both algorithm and required FEN. So far, we understand that moving from extreme points (EP-PM) to Pareto front (PF-PM) for training step increases the accuracy of prediction model. However, moving further and choosing only random points from evolver (RO-PM) is not the solution for covering all possible given COPs (extreme and random like instances). In other words, there should be a trade-off relation between moving towards random points from extreme and random instances in multi-objective evolver population. Therefore, our preference for training phase is a combination set of Pareto front and random instances from multi-objective evolver. Not only it covers instances that are hard/easy for one and easy/hard for the other algorithms, but also it can predict general given COPs more accurately.

To train our Pareto front with random instances prediction model (PFR-PM), we use 3000 points from evolver population set (1500 each). To compare and assess the prediction model accuracy we experiment our PFR-PM with the same testing COPs for the former models. The results for Pareto front and random prediction model (PFR-PM) are shown in Table \ref{table:paretrandom}. It is observed that including both Pareto front line and random only instances can be effective in accuracy improvement. Analysing the results, it is obvious the error rate for both predicted algorithm and required FEN are decreased. Also, the model is able to predict required FEN using other algorithms (not the best one) more accurate. This can be concluded by having lower error rates for FENs of other algorithms (not the best one) for PFR-PM. The reason behind this is that to cover all possible given COPs, we use both Pareto front and random points from evolver population for out training phase. In other words, our model is trained with constraint characteristics and features of both types of COPs (Pareto front and random instances).

It is shown that choosing the proper subsets for training phase is effective in prediction model accuracy. The results for four prediction models suggest that moving from extreme points to random instances can improve the quality of prediction model. It is found that there is a trade-off relationship in choosing instances that are close to random or extreme points (from evolver) for training phase. Results analysis shows by selecting a combination subsets from both random and Pareto front instances (PFR-PM) for training step, we can improve our model prediction quality. This improvement is in both selected algorithm and also its required FEN. By selecting the best model (PFR-PM), in the following, we examine it in a more detailed approach.

\section{Experiments on Benchmarks}
Our goal is to design highly accurate prediction model for a given COP based on its constraint features. As mentioned earlier, in order to improve the model accuracy, it needs to be trained with COP instances that are generated using multi-objective evolver. So far, we analysed the results for all four types of prediction models trained with extreme points (EP-PM), Pareto front (PF-PM), random only (RO-PM) and combination of Pareto front with random (PFR-PM) instances. We have experimented our prediction model with various subsets of training data to analyse the best preference. As results indicate, the most accurate prediction model is the one which is trained with combination of Pareto front and random subset of evolver population (PFR-PM). This model is capable of predicting algorithms for almost all possible given testing COPs such as the ones similar to extreme (hard/easy) or ordinary instances (random) in a multi-objective evolver population. Also, the prediction ability for required function evaluation number (FEN) has been significantly increased. Therefore, in order to assess our optimised prediction model (see Section \ref{Paretorandomsec}), we decide to experiment it  with our newly designed benchmark. In order to analyse the capability of the prediction model (PFR-PM) on constraints, we use fixed objective function with various numbers of linear, quadratic (and their combination) constraints. Then, we test other well-known objective function to see the relationship of constraints and our prediction model.

For this experiment, we train our PFR-PM with 3000 instances from both  Pareto front and random instances of evolving algorithm population set (1500 each). We also use optimised algorithm parameter settings for each algorithm suggested in \cite{takahama2010constrained,wang2009hybrid,arnold20121+}. In order to show the accuracy of model on prediction over constraints, we examine the model on various well-known objective functions such as Sphere (bowl-shaped), Ackley (many local optima) and Rosenbrock (valley-shaped). Also, to evaluate the effectiveness of our model on constrained problems, we use various numbers and types of constraints. Tables \ref{table:benchmarksphere}, \ref{table:benchmarkackley} and \ref{table:benchmarkrosenbrock} show the prediction results for Sphere, Ackley and Rosenbrock objective functions respectively. The results show the number of correct algorithm types prediction (success rate) from 30 different tests. Also, one step further, the average deviations of required FEN (the correct and predicted one) for the predicted algorithms are calculated.

Table \ref{table:benchmarksphere} compares the prediction results for our proposed model (PFR-PM) and random only model (RO-PM) for Sphere COPs. The results indicate the effectiveness of choosing the proper subset training instances. It is observed that the prediction algorithm success rate for our proposed model (PFR-PM) is significantly better than RO-PM for all Sphere COPs using various combinations of constraints. The success rate (out of 30 tests) for newly testing given COP is significantly higher for PFR-PM comparing to RO-PM. Also, the low value average deviation of predicted FEN and actual one for PFR-PM represents its higher accuracy in predicting the algorithm performance in terms of function evaluation number.

By observing the Tables \ref{table:benchmarkackley} and \ref{table:benchmarkrosenbrock}, we realise that our prediction model (PFR-PM) is reliable in predicting with only constraints. In other words, experimenting other types of objective functions (Bowl-shaped, many local-optima and valley shaped) with accurate results shows the ability of the model to predict based on constraints. Based on the Table \ref{table:benchmarkackley}, the lower value of FEN average deviation indicates the higher accuracy of PFR-PM for Ackley COPs. Also, the results for Rosenbrock COPs with 1 to 5 linear, quadratic constraints (and their combination) shows the accuracy of PFR-PM comparing to RO-PM. The average deviation of FEN for Rosenbrock problems denotes the significantly close predicted FEN with PFR-PM (see Table \ref{table:benchmarkrosenbrock}).

As mentioned before, the output of our proposed prediction model (PFR-PM) includes predicted algorithm with its required FEN. It is observed that the prediction model is capable of suggesting the best algorithm and required FEN based on constraint features of given COP. Due to the stochastic nature of evolutionary optimisation, the above benchmark tests are repeated 30 times and the two-tail t-test significance is performed for average deviations of FEN. The significant level $\alpha$ is considered as 0.05. The $p$-values for significance of a difference between FEN average deviation of Pareto front with random (PFR-PR) and random only (RO-PR) models for each Sphere, Ackley and Rosenbrock are shown in Tables \ref{table:benchmarksphere}, \ref{table:benchmarkackley} and \ref{table:benchmarkrosenbrock} respectively. The results show that the difference in FEN Average deviation are significant and less than 0.05.

As discussed earlier, the idea of designing a prediction model based on instance features is rather a novel approach in algorithm selection problem. Training a model with COP instances from multi-objective evolver improves the prediction accuracy.  The performance prediction (FEN) and suggested algorithm can be used to produce the final output of our prediction model. As we know selecting a suitable algorithm for a given problem requires substantial amount of time. In contrast, in our approach, we only need to extract features of a problem once and the model produces the final output. It is observed that selecting different sets of training instances improves the prediction model success rate. We designed and examined various prediction model using different subsets of problem instances from evolver population set. In order to show the ability of the prediction model only based on constraints features we use various objective functions. Results for these COPs with different combinations of objective functions and constraints indicate that the model is highly accurate in algorithm and required FEN prediction.
	
\section{Conclusion}
In this study, we examined the impact of  different types of problem instances that can be used in prediction models for constrained continuous optimisation. 

Our resulting prediction model captures the links between constraint features, algorithm type performance and the required function evaluation number. The model inputs are considered as constraint features and selected parameter settings. The outputs includes the required function evaluation number and most suited algorithm type to solve the given COP. 

The model was trained (using NN learning strategy) with evolved COP instances. To improve the accuracy of the model we used evolved instances that are hard/easy for one and easy/hard for the other algorithms. These training instances are generated with multi-objective evolver. We first, chose various subsets of instances from multi-objective evolver population set. It is observed that the a model using combination of Pareto front line and random points from population set has the highest accuracy in predicting best algorithm types for a given COP. We then, tested our prediction model with different objective functions and constraints types. The results indicate our prediction model is reliable to suggest and predict most suited algorithm and required FEN using problem's constraint features. Our approach shows the relationship between constraint features and various algorithm performances. The results clearly demonstrate the ability of prediction model to predict the algorithm and required FEN using only constraint features.

\begin{table}	  	
\vspace{-3cm}	
\caption {Predicted and actual most suited algorithm type/required FEN for Sphere function with linear constraint(s). The prediction model is trained with extreme points (EP-PM) from multi objective evolver population. DE hard/easy (1 C) is a problem instance that is hard/easy for DE algorithm but easy/hard for the others.}
{%
\hspace{-4cm}
	\begin{tabular}{|p{2.1cm}|p{1.2cm}|p{0.8cm}|p{0.6cm}|p{1.2cm}|p{1cm}|p{1.0cm}|p{1.2cm}|p{0.9cm}|p{1.0cm}|p{1.2cm}|p{1.2cm}|p{1.0cm}|}
		
\hline \hline
Instances name	& Predicted alg. & Actual alg. & Error & Predicted FEN for DE & Actual FEN for DE & Error for DE &Predicted FEN for ES & Actual FEN for ES& Error for ES & Predicted FEN for PSO & Actual FEN for PSO & Error for PSO  \\ \hline
DE hard (1 c)&  ES & ES & NO & 83.2K & 86.3K & -3.1K & \textbf{43.7K}  & \textbf{41.5K} & +2.2K & 46.8K & 43.2K &  +3.6K \\ \hline
ES hard (1 c)&  PSO & DE & YES & 43.4K & \textbf{45.7K} & -2.3K  & 80.9K & 84.2K &  -3.3K & \textbf{41.6K} & 48.3K& -6.7K\\ \hline
PSO hard (1 c)&  DE & DE & NO & \textbf{38.9K} & \textbf{37.2K} & +1.7K & 43.2K & 41.8K & +1.4K &76.2K & 80.1K & -3.9K \\ \hline
DE hard (2 c)&  ES & ES & NO & 83.5K & 87.4K & -3.9K & \textbf{42.5K} & \textbf{45.2K} & -2.7K & 44.2K & 43.6K & +1.4K\\ \hline
ES hard (2 c)&  DE & DE & NO & \textbf{47.2K} & \textbf{46.4K} & +0.8K & 84.3K & 88.3K & -4.0K & 49.4K& 46.8K& +2.6K\\ \hline
PSO hard (2 c)& DE  & DE & NO & \textbf{41.5K} & \textbf{43.6K} & -2.1K & 46.3K & 45.1K &+1.2K & 85.5K & 83.2K &+2.2K \\ \hline
DE hard (3 c)& ES  & ES & NO & 87.3K & 92.4K & -5.1K & \textbf{43.8K} & \textbf{45.2K} & -1.4K & 45.6K & 47.2K & -1.6K\\ \hline
ES hard (3 c)&  PSO & PSO & NO  & 45.6K &  51.2K  & -5.6K &95.5K & 91.2K&+4.3K & \textbf{43.5K} & \textbf{48.2K} & -4.7K\\ \hline
PSO hard (3 c)& DE  & DE & NO & \textbf{47.2K} & \textbf{49.6K} & -2.4K & 51.2K & 53.5K & -2.3K  & 92.1K& 89.5K & +2.6K\\ \hline
DE hard (4 c)&  PSO & ES  & YES  & 95.2K & 93.7K & -1.5K &52.4K & \textbf{47.2K} & +5.2K & \textbf{49.2K} & 50.2K & -1.0K\\ \hline
ES hard (4 c)&  PSO & PSO & NO  & 51.3K & 49.2K & +2.1K & 85.3K & 89.1K & -3.8K & \textbf{46.3K} & \textbf{48.9K} & -2.6K\\ \hline
PSO hard (4 c)& DE  & ES  & YES & \textbf{49.2K} & 52.7K & -3.5K & 53.4K & \textbf{50.8K} & +2.6K & 89.3K & 87.5K & +1.8K\\ \hline
DE hard (5 c)& ES  & ES & NO & 94.7K  & 96.3K & -1.6K & \textbf{49.3K} & \textbf{53.3K} & -4.0K & 57.1K & 55.3K  & +1.8K \\ \hline
ES hard (5 c)&  DE & DE & NO & \textbf{51.4K} & \textbf{53.8K} & -2.4K & 94.2K & 92.6K& +1.6K & 55.2K & 54.7K & -2.5K\\ \hline
PSO hard (5 c)& DE  & DE & NO & \textbf{49.1K} & \textbf{51.3K} & -2.2K & 57.9K & 55.2K & +2.7K & 89.5K & 93.2K & -3.7K\\ \hline \hline

DE easy (1 c)& DE  & DE & NO & \textbf{45.3K} & \textbf{48.9K} & -3.6K & 81.5K & 85.2K& -3.7K &75.4K &79.1K & -3.7K\\ \hline
ES easy (1 c)& ES & ES & NO & 87.3K & 81.4K & +5.9K & \textbf{48.3K} & \textbf{54.7K} & -6.4K & 79.4K & 81.7K & -2.3K\\ \hline
PSO easy (1 c)& PSO  & PSO & NO & 92.5K & 87.1K & +5.4K & 81.4K & 84.2K& -2.8K &\textbf{48.2K}&\textbf{41.9K} &+6.3K\\ \hline
DE easy (2 c)& DE  & DE & NO & \textbf{49.4K} & \textbf{44.5K} & +4.9K & 89.3K & 83.1K & +6.2K &78.9K & 81.6K & -2.7K\\ \hline
ES easy (2 c)&  ES & ES & NO & 78.4K & 83.6K & -5.2K& \textbf{51.4K} & \textbf{55.3K}&-3.9K &79.3K & 78.9K & +0.4K\\ \hline
PSO easy (2 c)& PSO  & PSO & NO & 84.2K & 81.4K & +2.8K &78.3K  &83.9K & -5.6K & \textbf{51.6K} & \textbf{49.4K} & 2.2K \\ \hline
DE easy (3 c)&  DE & DE & NO & \textbf{41.8K} & \textbf{48.4K} & -6.6K  & 85.8K & 89.4K & -3.6K &86.3K & 84.5K & +1.8K\\ \hline
ES easy (3 c)&  ES & ES & NO & 91.4K & 87.8K & +3.6K & \textbf{48.2K} & \textbf{46.2K} & +2.0K & 51.3K & 48.0K &+3.3K\\ \hline
PSO easy (3 c)&  PSO & PSO & NO & 88.4K  & 89.5K  & -1.1K & 93.2K & 87.3K  & +5.9K & \textbf{46.8K}& \textbf{49.1K} & -2.3K\\ \hline
DE easy (4 c)&  DE & DE & NO & \textbf{51.6K}  & \textbf{53.2K} & -1.6K & 83.5K &85.1K & -1.6K &91.4K & 93.8K& -2.4K \\ \hline
ES easy (4 c)&  ES & ES & NO & 85.3K & 89.4K &  -4.1K & \textbf{51.2K} & \textbf{48.9K}& +2.3K &89.4K & 87.2K& +2.2K\\ \hline
PSO easy (4 c)& PSO  & DE & YES &  57.2K & \textbf{55.5K} & +1.7K& 84.1K & 92.4K& -8.3K & \textbf{54.2K} & 68.9K& -14.7K\\ \hline
DE easy (5 c)&  DE & DE & NO & \textbf{59.2K} & \textbf{57.8K} & +1.4K & 93.2K & 95.3K & -2.1K &81.3K & 81.5K& -0.2K\\ \hline
ES easy (5 c)& ES  & ES & NO &  88.3K& 91.9K & -3.6K &\textbf{53.2K}  & \textbf{55.7K}& -2.5K & 86.3K & 81.1K & 5.2K \\ \hline
PSO easy (5 c)& PSO  & PSO & NO & 89.3K & 91.3K & -2.0K & 84.3K & 82.9K& +1.4K & \textbf{55.8K}& \textbf{57.6K} & -1.8K\\ \hline \hline

Rnd. 1 (1 c)	& DE  & PSO & YES & \textbf{53.2K} & 65.7K & -12.5K  & 65.3K  & 56.9K &  +8.4K & 62.4K &  \textbf{53.2K} & +9.2K\\ \hline
Rnd. 2 (1 c)	& ES  & DE & YES & 77.2K &\textbf{61.9K} & 15.3K & \textbf{43.5K} & 64.2K & -20.7 & 51.9K & 65.2K & -13.3K\\ \hline
Rnd. 3	(2 c) & ES  & DE & YES & 71.4K & \textbf{59.8K}  & +11.6K  & \textbf{48.1K} & 65.3K &  -17.2K & 60.2K & 69.6K & -9.4K\\ \hline
Rnd. 4 (2 c)  & PSO	& ES  & YES & 61.2K & 72.2K & -11.0K & 68.8K & \textbf{59.4K} & +9.4K & \textbf{55.3K} & 71.4K & -16.1K\\ \hline
Rnd. 5 (3 c) & DE  & DE & NO & \textbf{45.2K}  & \textbf{65.4K} & -20.2K & 77.2K & 66.9K & +10.3K & 64.5K & 74.6K &-10.1K \\ \hline
Rnd. 6 (3 c) &  ES & PSO & YES  & 71.9K & 82.4K  & -10.5K & \textbf{45.6K} & 78.3K & -32.7K & 50.3K & \textbf{61.4K}&-11.1K \\ \hline
Rnd. 7 (4 c) & DE  & ES & YES & \textbf{71.6K} & 83.2K & -11.6K & 80.3K & \textbf{65.8K} & +14.5K & 89.2K & 78.9K & +10.3K \\ \hline
Rnd. 8 (4 c) & PSO  & DE & YES & 84.7K & \textbf{71.1K} & +13.6K & 68.8K & 79.2K & -10.4K & \textbf{62.7K} & 73.2K & -10.5K\\ \hline
Rnd. 9 (5 c) & ES  & DE & YES & 91.8K & \textbf{82.7K} & +9.1K  & \textbf{83.9K}  & 93.8K & -9.9K & 89.0K & 97.3K & -8.3K\\ \hline
Rnd. 10 (5 c)& DE  & PSO & YES & \textbf{78.4K}  & 93.2K  & -14.8K & 79.2K & 89.5K& -10.3K & 89.3K & \textbf{68.4K}& +20.9K\\ \hline
Rnd. 11 (1 c)	&  ES & DE & YES & 56.3K & \textbf{44.6K} & +11.7K & \textbf{49.3K} & 59.2K& -9.9K & 65.2K&  64.2K& -8.0K\\ \hline
Rnd. 12 (1 c) & ES	& DE  & YES & 54.2K  &\textbf{45.2K}  & +9.0K &\textbf{48.2K} & 58.9K & -10.7K & 61.2K & 69.4K & -8.2K\\ \hline
Rnd. 13 (2 c) &DE& PSO & YES & \textbf{48.2K}  & 61.2K  & -13.0K & 78.5K & 64.3K &  +14.2K  & 50.9K & \textbf{59.1K} & -8.2K \\ \hline
Rnd. 14 (2 c) &ES	&  PSO & YES&70.2K & 79.0K & -8.8K & \textbf{61.2K} & 75.3K & -14.1K  & 71.2K& \textbf{63.4K} & +7.8K \\ \hline
Rnd. 15 (3 c) &  DE & PSO & YES & \textbf{63.4K} & 71.9K  & -8.5K & 71.3K & 79.3K& -8.0K & 73.1K& \textbf{65.3K} &+7.8K\\ \hline
Rnd. 16 (3 c) &  PSO & DE & YES & 77.5K & \textbf{69.2K} & +8.3K & 89.5K & 79.3K&+10.2K & \textbf{74.3K} & 70.1K & +6.2K\\ \hline
Rnd. 17 (4 c) &  ES & ES & NO & 69.3K & 73.9K & -4.6K & \textbf{68.5K}  & \textbf{75.9K} & -7.4K & 69.K & 77.4K& -7.7K\\ \hline
Rnd. 18 (4 c) &  DE & PSO & YES & \textbf{71.4K} & 85.3K & -13.9K & 81.2K & 78.3K& +2.9K & 75.3K &\textbf{68.3K} & +7.0K\\ \hline
Rnd. 19 (5 c) &  DE & ES & YES & \textbf{81.2K} & 93.9K & -12.7K & 83.2K & \textbf{91.2K} & -8.0K &87.2K & 95.9K& -8.7K\\ \hline
Rnd. 20 (5 c)&  PSO & PSO & NO & 81.2K & 90.9K & -9.7K & 81.3K & 93.2K& -11.9K & \textbf{78.2K} & \textbf{87.3K} &-9.1K\\ \hline\hline		

\end{tabular}}
\setlength{\tabcolsep}{5em}

\label{table:extonlylin}
	
\end{table}

\begin{table}	  
\vspace{-3cm}	
\caption {Predicted and actual most suited algorithm type/required FEN for Sphere function with linear constraint(s). The prediction model is trained with pareto front line (PF-PM) from multi objective evolver population. DE hard/easy (1 C) is a problem instance that is hard/easy for DE algorithm but easy/hard for the others.}
{%
\hspace{-4cm}
	\begin{tabular}{|p{2.1cm}|p{1.2cm}|p{0.8cm}|p{0.6cm}|p{1.2cm}|p{1cm}|p{1.0cm}|p{1.2cm}|p{0.9cm}|p{1.0cm}|p{1.2cm}|p{1.2cm}|p{1.0cm}|}
		
		\hline \hline
		Instances name	& Predicted alg. & Actual alg. & Error & Predicted FEN for DE & Actual FEN for DE & Error for DE &Predicted FEN for ES & Actual FEN for ES& Error for ES & Predicted FEN for PSO & Actual FEN for PSO & Error for PSO  \\ \hline
		
		DE hard (1 c)& ES  & ES & NO  & 82.7K & 86.3K & -3.6& \textbf{43.5K}  & \textbf{41.5K} & 2.0K & 47.3K & 43.2K & 4.1K  \\ \hline
		ES hard (1 c)&  PSO & DE & YES &  42.3K & \textbf{45.7K} & -3.4K  & 86.4K & 84.2K & 2.2K  & \textbf{40.3K} & 48.3K& -8.0K\\ \hline
		PSO hard (1 c)&  DE & DE & NO & \textbf{35.1K} & \textbf{37.2K} & -2.1K & 43.8K & 41.8K & 2.0K & 77.4K & 80.1K &  -2.7K\\ \hline
		DE hard (2 c)&  ES & ES & NO & 89.4K & 87.4K & 2.0K  & \textbf{43.1K} & \textbf{45.2K} & -2.1K & 45.7K & 43.6K &2.1K \\ \hline
		ES hard (2 c)& DE & DE & NO & \textbf{48.5K} & \textbf{46.4K} & 2.1K &86.9K & 88.3K & -1.4K & 48.2K & 46.8K& 1.4K\\ \hline
		PSO hard (2 c)& DE  & DE & NO & \textbf{42.3K} & \textbf{43.6K} & -1.3K & 47.2K  & 45.1K & 2.1K & 84.6K & 83.2K & 1.4K\\ \hline
		DE hard (3 c)&  ES & ES & NO & 87.6K & 92.4K & -4.8K & \textbf{42.9K}  & \textbf{45.2K} & -2.3K & 46.2K & 47.2K & -1.0\\ \hline
		ES hard (3 c)& DE  & PSO & YES  & \textbf{49.3K} &  51.2K  & -1.9K & 94.5K & 91.2K& 3.3K & 50.7K & \textbf{48.2K} & 2.5K\\ \hline
		PSO hard (3 c)& DE  & DE & NO & \textbf{47.5K} & \textbf{49.6K} & -2.1K  & 55.7K & 53.5K & 2.2K  & 93.1K & 89.5K & 3.6K \\ \hline
		DE hard (4 c)& ES  & ES  & NO  & 94.9K & 93.7K & 1.2K & \textbf{49.2K} & \textbf{47.2K} & 2.0K & 53.6K & 50.2K & 3.4K\\ \hline
		ES hard (4 c)& PSO & PSO & NO  & 52.4K & 49.2K & 3.2K & 86.9K & 89.1K & -2.2K & \textbf{45.1K} & \textbf{48.9K} & -3.8K \\ \hline
		PSO hard (4 c)& ES  & ES & NO & 55.1K & 52.7K & 2.4K & \textbf{51.9K} & \textbf{50.8K} & 1.1K  & 85.2K & 87.5K & -2.3K\\ \hline
		DE hard (5 c)& ES  & ES & NO & 94.1K  & 96.3K & -2.2 & \textbf{55.9K} & \textbf{53.3K} & 2.6K & 58.3K & 55.3K  & 3.0K \\ \hline
		ES hard (5 c)&  DE & DE & NO & \textbf{51.9K} & \textbf{53.8K} & -1.9K & 95.1K & 92.6K& 2.5K  & 52.9K & 54.7K & -1.8K\\ \hline
		PSO hard (5 c)& ES  & DE & YES & 54.7K & \textbf{51.3K} & 3.4K & \textbf{53.1K} & 55.2K & -2.1K& 89.4K & 93.2K & -3.8K\\ \hline \hline
		
		DE easy (1 c)& PSO  & DE & YES & 72.3K & \textbf{48.9K} & 23.4K & 80.4K & 85.2K& -4.8K & \textbf{71.2K} &79.1K & -7.9K \\ \hline
		ES easy (1 c)& ES & ES & NO & 78.4K & 81.4K & -3.0K & \textbf{47.2K} & \textbf{54.7K} & -7.5K & 78.9K & 81.7K & -2.8K\\ \hline
		PSO easy (1 c)& PSO  & PSO & NO & 91.4K & 87.1K & 4.3K & 80.7K & 84.2K& -3.5K & 45.8K &\textbf{41.9K} & 3.9K\\ \hline
		DE easy (2 c)&  DE& DE & NO & \textbf{48.1K} & \textbf{44.5K} & 3.6K & 85.9K & 83.1K & 2.8K & 85.1K & 81.6K & 3.5K\\ \hline
		ES easy (2 c)& ES & ES & NO & 79.4K & 83.6K & -4.2K & \textbf{52.9K} & \textbf{55.3K}& -2.4K & 80.7K & 78.9K & 1.8K\\ \hline
		PSO easy (2 c)& PSO  & PSO & NO & 83.5K & 81.4K & 2.1K& 79.3K &83.9K & -4.6K & \textbf{47.1K} & \textbf{49.4K} & -2.3K \\ \hline
		DE easy (3 c)&  DE & DE & NO & \textbf{42.4K} & \textbf{48.4K} &  -6.0K & 86.2K & 89.4K & -3.2K & 87.5K & 84.5K & 3.0K\\ \hline
		ES easy (3 c)&  ES & ES & NO & 90.4K & 87.8K & 2.6K &  \textbf{47.9K} & \textbf{46.2K} & 1.7K & 50.5K & 48.0K &2.5K\\ \hline
		PSO easy (3 c)& PSO  & PSO & NO & 86.4K  & 89.5K  & -3.1K & 94.1K & 87.3K  & 6.8K & \textbf{47.9K}& \textbf{49.1K} & -1.2K\\ \hline
		DE easy (4 c)& DE  & DE & NO & \textbf{50.8K}  & \textbf{53.2K} & -2.4K & 84.2K &85.1K & -0.9K & 95.3K & 93.8K& 1.5K \\ \hline
		ES easy (4 c)& ES  & ES & NO & 86.3K & 89.4K & -3.1K  & \textbf{46.2K} & \textbf{48.9K}& -2.7K  & 90.3K & 87.2K& 3.1K\\ \hline
		PSO easy (4 c)& PSO  & DE & YES &   60.4K & \textbf{55.5K} & 4.9K & 88.2K & 92.4K& -4.2K  &  \textbf{58.8K} & 68.9K& -10.1K \\ \hline
		DE easy (5 c)&  DE & DE & NO & \textbf{56.7K} & \textbf{57.8K} & -1.1K & 92.5K & 95.3K & -2.7K & 83.2K & 81.5K& 1.7K\\ \hline
		ES easy (5 c)& ES  & ES & NO  & 87.3K & 91.9K & -4.6K & \textbf{52.8K} & \textbf{55.7K}& -2.9K &85.4K & 81.1K & 4.3K\\ \hline
		PSO easy (5 c)& PSO  & PSO & NO & 88.2K & 91.3K & -3.1K & 85.2K & 82.9K& 2.3K & \textbf{56.2K} & \textbf{57.6K} & -1.4K \\ \hline \hline

		Rnd. 1 (1 c)	& PSO  & PSO & NO & 60.8K & 65.7K & -4.9K  & 60.2K  & 56.9K & +3.3K  & \textbf{48.8K} &  \textbf{53.2K} & -4.4K\\ \hline
		Rnd. 2 (1 c)	& DE  & DE & NO & \textbf{67.3K} &\textbf{61.9K} & +5.4K & 68.1K & 64.2K & +3.9K  & 71.3K & 65.2K & +6.1K \\ \hline
		Rnd. 3	(2 c) & DE  & DE & NO & \textbf{51.5K} & \textbf{59.8K}  & -8.3K  & 59.9K & 65.3K & -5.2K  & 63.8K & 69.6K & -5.8K \\ \hline
		Rnd. 4 (2 c)  & ES	& ES  & NO & 65.9K & 72.2K & -6.3K & \textbf{64.1K} & \textbf{59.4K} & +4.7K & 64.7K & 71.4K & -6.7K \\ \hline
		Rnd. 5 (3 c) & ES  & DE & YES & 59.2K  & \textbf{65.4K} & -6.2K & \textbf{57.1K} & 66.9K & -9.8K & 67.3K & 74.6K &  -7.3K \\ \hline
		Rnd. 6 (3 c) &  PSO & PSO & NO  & 74.9K & 82.4K  & -7.5K & 69.2K & 78.3K & -9.1K & \textbf{55.1K} & \textbf{61.4K}& -6.3K \\ \hline
		Rnd. 7 (4 c) & ES  & ES & NO & 87.6K & 83.2K & +4.4K & \textbf{71.3K} & \textbf{65.8K} & +5.5K & 72.1K & 78.9K & -6.8K\\ \hline
		Rnd. 8 (4 c) & ES  & DE & YES & 78.3K & \textbf{71.1K} & +7.2K &\textbf{70.5K} & 79.2K & -8.7K & 77.9K & 73.2K & +4.7K\\ \hline
		Rnd. 9 (5 c) & DE  & DE & NO & \textbf{87.9K} & \textbf{82.7K} & +5.2K  & 88.2K & 93.8K & -5.6K & 91.9K & 97.3K & -5.4K\\ \hline
		Rnd. 10 (5 c)& PSO  & PSO & NO & 79.1K & 93.2K  & -14.1K & 85.6K & 89.5K& -3.9K & \textbf{75.6K} & \textbf{68.4K}& +7.2K\\ \hline\hline	
		
		Rnd. 11 (1 c)	&  DE & DE & NO & \textbf{51.2K} & \textbf{44.6K} & +6.6K & 52.1K & 59.2K& -7.1K & 57.1K&  64.2K& -7.1K \\ \hline
		Rnd. 12 (1 c) & DE	& DE  & NO & \textbf{49.2K}  &\textbf{45.2K}  & +4.0K &52.1K & 58.9K & -6.8K & 62.4K & 69.4K & -7.0K \\ \hline
		Rnd. 13 (2 c) &PSO& PSO & NO & 53.2K & 61.2K  & -8.0K & 60.5K & 64.3K & -3.8K  & \textbf{52.7K} & \textbf{59.1K} & -6.4K \\ \hline
		Rnd. 14 (2 c) & PSO	&  PSO & NO &72.9K & 79.0K & -6.1K & 65.9K & 75.3K & -9.4K   & \textbf{59.6K}& \textbf{63.4K} & -3.8K  \\ \hline
		Rnd. 15 (3 c) &  ES & PSO & YES & 65.1K & 71.9K  & -6.8K & \textbf{59.9K} & 79.3K& -19.4K & 60.9K& \textbf{65.3K} & -4.4K\\ \hline
		Rnd. 16 (3 c) &  DE & DE & NO & \textbf{63.1K} & \textbf{69.2K} & -7.9K & 72.5K & 79.3K& -6.8K &74.2K & 70.1K & 6.1K\\ \hline
		Rnd. 17 (4 c) &  ES & ES & NO & 69.9K & 73.9K & -4.0K & \textbf{68.4K}  & \textbf{75.9K}& -7.5K & 70.3K & 77.4K& -7.1K \\ \hline
		Rnd. 18 (4 c) &  PSO & PSO & NO & 74.8K & 85.3K & -10.5K & 84.0K & 78.3K& +5.7K & \textbf{72.6K} &\textbf{68.3K} & +4.3K \\ \hline
		Rnd. 19 (5 c) &  DE & ES & YES & \textbf{85.1K} & 93.9K & -8.1K & \textbf{85.8K} & 91.2K & -6.1K &98.9K & 95.9K& +3.0K\\ \hline
		Rnd. 20 (5 c)&  PSO & PSO & NO & 85.1K & 90.9K & -5.8K & 87.9K & 93.2K& -5.3K  & \textbf{82.5K} & \textbf{87.3K} & -4.8K\\ \hline\hline	
	\end{tabular}}
	\setlength{\tabcolsep}{5em}
	
	\label{table:Paretoonly}
	
\end{table}

\begin{table}	
\vspace{-3cm}	
\caption {Predicted and actual most suited algorithm type/required FEN for Sphere function with linear constraint(s). The prediction model is trained with random points (RO-PM) from multi objective evolver population. DE hard/easy (1 C) is a problem instance that is hard/easy for DE algorithm but easy/hard for the others.}
{%
\hspace{-4cm}
	\begin{tabular}{|p{2.1cm}|p{1.2cm}|p{0.8cm}|p{0.6cm}|p{1.2cm}|p{1cm}|p{1.0cm}|p{1.2cm}|p{0.9cm}|p{1.0cm}|p{1.2cm}|p{1.2cm}|p{1.0cm}|}
		
		\hline \hline
		Instances name	& Predicted alg. & Actual alg. & Error & Predicted FEN for DE & Actual FEN for DE & Error for DE &Predicted FEN for ES & Actual FEN for ES& Error for ES & Predicted FEN for PSO & Actual FEN for PSO & Error for PSO  \\ \hline
		
		DE hard (1 c)& PSO  & ES & YES & 93.6K & 86.3K & +7.3K & 53.5K  & \textbf{41.5K} & +12K & \textbf{38.1K} & 43.2K & -5.1K  \\ \hline
		ES hard (1 c)&  DE & DE & NO & \textbf{51.8K} & \textbf{45.7K} & +6.1K  & 78.1K & 84.2K &  -6.1K & 53.7K & 48.3K& +5.4K \\ \hline
		PSO hard (1 c)& PSO  & DE & YES & 58.8K & \textbf{37.2K} & +21.6K & 73.5K & 41.8K & +31.7K  & \textbf{57.3K} & 80.1K & -22.8K \\ \hline
		DE hard (2 c)&  ES & ES & NO & 66.8K & 87.4K & -20.6K & \textbf{57.9K} & \textbf{45.2K} & +12.7K & 72.8K & 43.6K & +29.2K \\ \hline
		ES hard (2 c)&  PSO & DE &YES  & 67.2K & \textbf{46.4K} & +20.8K  & 73.2K & 88.3K &  -15.2K & \textbf{53.7K} & 46.8K& +6.9K\\ \hline
		PSO hard (2 c)& DE   & DE & NO & \textbf{55.7K} & \textbf{43.6K} & +12.1K & 73.6K & 45.1K & +28.5K & 77.8K & 83.2K & -5.4K\\ \hline
		DE hard (3 c)&  PSO & ES & YES & 73.4K & 92.4K & -19.0K & 59.2K & \textbf{45.2K} & +14.0K & \textbf{57.3K} & 47.2K & +10.1K\\ \hline
		ES hard (3 c)&  DE & PSO & YES  & \textbf{62.7K} &  51.2K  & +11.5K & 73.6K & 91.2K& -17.6K & 64.3K & \textbf{48.2K} & +16.1K\\ \hline
		PSO hard (3 c)& ES  & DE &YES  & 59.2K & \textbf{49.6K} & +9.6K & \textbf{43.2K} & 53.5K & -10.3K  & 103.6K & 89.5K & +14.1K\\ \hline
		DE hard (4 c)&  PSO & ES  &  YES & 88.2K & 93.7K & -5.5K  & 51.2K & \textbf{47.2K} & +4.0K & \textbf{49.1K} & 50.2K & -1.1K\\ \hline
		ES hard (4 c)&  ES & PSO & YES  & \textbf{39.8K} & 49.2K & -9.4K  & 96.3K & 89.1K & +7.2K & 41.2K & \textbf{48.9K} & -7.7K\\ \hline
		PSO hard (4 c)& DE  & ES  &YES  & \textbf{63.9K} & 52.7K & +11.2K & 64.6K & \textbf{50.8K} & +13.8K & 83.8K & 87.5K & -3.7K\\ \hline
		DE hard (5 c)&  PSO & ES & YES & 102.6K  & 96.3K & +6.3K & 62.3K & \textbf{53.3K} & +9.0K  & \textbf{49.2K} & 55.3K  & -6.1K \\ \hline
		ES hard (5 c)& PSO  & DE &YES  & 47.9K & \textbf{53.8K} & -5.9K & 84.8K & 92.6K& -7.8K & \textbf{47.1K} & 54.7K & -7.6K\\ \hline
		PSO hard (5 c)& ES  & DE &YES  & 59.9K & \textbf{51.3K} & +8.6K  & \textbf{45.2K} & 55.2K & -10.0K & 82.7K & 93.2K & -10.5K \\ \hline \hline

		DE easy (1 c)& DE  & DE & NO & \textbf{41.2K} & \textbf{48.9K} &-7.7K & 78.2K & 85.2K& -7K & 67.2K &79.1K &-11.9K \\ \hline
		ES easy (1 c)& ES & ES & NO & 68.2K & 81.4K & -13.2K & \textbf{64.2K} & \textbf{54.7K} & 9.6K & 72.9K & 81.7K & -8.8K \\ \hline
		PSO easy (1 c)& ES  & PSO & YES & 81.2K & 87.1K & 4.1K  & \textbf{62.1K} & 84.2K& -22.1K & 63.4K &\textbf{41.9K} & 21.5K\\ \hline
		DE easy (2 c)&  DE & DE & NO & \textbf{54.2K} & \textbf{44.5K} & 9.7K & 68.4K & 83.1K & -14.7K  & 71.4K & 81.6K & -10.2K \\ \hline
		ES easy (2 c)&  PSO & ES & YES  & 75.3K  & 83.6K & -8.3K & 69.2K & \textbf{55.3K}&13.9K &\textbf{65.1K} & 78.9K & -13.8K\\ \hline
		PSO easy (2 c)& PSO  & PSO & YES  & 68.3K & 81.4K &-13.1K & 94.1K &83.9K & 10.2K & \textbf{58.1K} & \textbf{49.4K} & 8.7K \\ \hline
		DE easy (3 c)&  DE & DE & NO & \textbf{59.1K} & \textbf{48.4K} & 10.7K  & 78.2K & 89.4K & -11.2K & 80.4K & 84.5K & -4.1K \\ \hline
		ES easy (3 c)&  PSO & ES & YES & 73.9K & 87.8K & -13.9K & 55.2K & \textbf{46.2K} & 9K & \textbf{42.4K} & 48.0K & -5.6K\\ \hline
		PSO easy (3 c)& PSO & PSO & NO  & 69.2K  & 89.5K  & -20.3K & 76.4K  & 87.3K  & -10.9K & \textbf{58.2K} & \textbf{49.1K} &9.1K \\ \hline
		DE easy (4 c)& DE & DE & NO & \textbf{64.2K} & \textbf{53.2K} & 11K  & 78.3K &85.1K & -6.8K & 87.3K & 93.8K&  -6.5K\\ \hline
		ES easy (4 c)&  ES& ES & NO & \textbf{78.3K} & 89.4K &-11.1K  & 59.1K & \textbf{48.9K}& 10.2K   & 82.5K & 87.2K& -4.7K\\ \hline
		PSO easy (4 c)& PSO  & DE & YES &   61.3K & \textbf{55.5K} & 5.8K & 85.9K & 92.4K& -6.5K   &  \textbf{60.4K} & 68.9K& -8.5K \\ \hline
		DE easy (5 c)&  PSO & DE & YES & 71.4K & \textbf{57.8K} &13.6K  & 99.1K & 95.3K & 3.8K &\textbf{70.9K} & 81.5K& -10.6K\\ \hline
		ES easy (5 c)&   ES & ES & NO & 86.9K & 91.9K & -5.0K & 62.3K & \textbf{55.7K}& 6.6K & 78.9K & 81.1K & -2.2K \\ \hline
		PSO easy (5 c)& PSO  & PSO & NO & 84.1K  & 91.3K & -7.2K & 76.9K & 82.9K& -6.0K & 63.4K & \textbf{57.6K} & 5.8K \\ \hline \hline

		Rnd. 1 (1 c)	&  PSO & PSO& NO  & 59.1K & 65.7K& -6.6K  &  51.3K & 56.9K & -5.6K  & \textbf{50.9K} & \textbf{53.2K} & -2.3K\\ \hline
		Rnd. 2 (1 c)	& DE  & DE & NO & \textbf{55.2K} & \textbf{61.9K} & -6.7K  &  67.4K & 64.2K & +3.2K  & 69.8K & 65.2K & +4.6K \\ \hline
		Rnd. 3 (2 c)	& DE & DE  & NO  & \textbf{49.2K} & \textbf{59.8K} &  -10.6K &  57.9K & 65.3K & -7.4K  & 73.4K & 69.6K & +3.8K\\ \hline
		Rnd. 4 (2 c)	&  ES & ES& NO & 64.1K & 72.2K&  -8.1K &  \textbf{63.2K} & \textbf{59.4K} & +3.8  & 65.3K & 71.4K & -6.1K \\ \hline
		Rnd. 5 (3 c)	&  ES & DE &YES  & 68.3K & \textbf{65.4K} & +2.9K  & \textbf{63.5K}  & 66.9K & -3.4K  & 69.9K & 74.6K & -4.7K \\ \hline
		Rnd. 6 (3 c)	&  PSO & PSO & NO & 73.9K & 82.4K& -8.5K  &  66.2K & 78.3K & -12.1K  & \textbf{56.8K} & \textbf{61.4K} & -4.6K \\ \hline
		Rnd. 7 (4 c)	& PSO  & ES & YES & 78.3K & 83.2K& -4.9K  & 73.5K  & \textbf{65.8K} & +7.7K  & \textbf{71.2K}  & 78.9K & -7.7K \\ \hline
		Rnd. 8 (4 c)	&  ES & DE & YES  & 76.9K & \textbf{71.1K}& +5.8K  &  \textbf{68.3K} & 79.2K & -10.9K  & 76.3K  & 73.2K & +3.1K\\ \hline
		Rnd. 9 (5 c)	&  DE & DE & NO & \textbf{88.3K} & \textbf{82.7K}& +5.6K   & 90.9K  & 93.8K &  -2.9K & 90.2K & 97.3K & -7.1K\\ \hline
		Rnd. 10 (5 c)	& PSO  & PSO & NO & 80.8K & 93.2K& -12.4K  &  83.5K & 89.5K & -6K  & \textbf{75.3K} & \textbf{68.4K} & +6.9K \\ \hline \hline
		
		Rnd. 11 (1 c)	& ES  & DE  & YES & 52.9K  & \textbf{44.6K}& +8.3K  & \textbf{52.4K}  & 59.2K & -6.8K  & 54.3K & 64.2K & -9.9K\\ \hline
		Rnd. 12 (1 c)	& DE  & DE & NO & \textbf{50.2K} & \textbf{45.2K}& +5K  & 51.2K  & 58.9K & -7.7K  & 62.5K & 69.4K & -6.9K \\ \hline
		Rnd. 13 (2 c)	&  PSO & PSO & NO & 67.3K & 61.2K&  6.1K & 60.8K  & 64.3K & -3.5K  & \textbf{49.2K}  & \textbf{59.1K} & -9.9K\\ \hline
		Rnd. 14 (2 c)	& PSO  & PSO &NO & 70.3K & 79.0K& -8.7K  & 68.3K  & 75.3K &  -7K & \textbf{57.9K} & \textbf{63.4K} & -5.5K\\ \hline
		Rnd. 15 (3 c)	& ES & PSO & YES & 66.9K & 71.9K&  -5K &  \textbf{55.2K} & 79.3K & -24.1K  & 58.3K & \textbf{65.3K} & -7K \\ \hline
		Rnd. 16 (3 c)	& DE  & DE & NO & \textbf{61.2K} & \textbf{69.2K}& -8K  &  70.5K & 79.3K & -8.8K  & 73.9K & 70.1K & 3.8K \\ \hline
		Rnd. 17 (4 c)	& ES & ES & NO & 67.9K & 73.9K&  -6K &  \textbf{65.2K} & \textbf{75.9K} & -10.7K  & 69.2K & 77.4K & -8.2K \\ \hline
		Rnd. 18 (4 c)	& PSO  & PSO & NO  & 72.7K & 85.3K& -12.6K  &  80.3K & 78.3K & +2K  & \textbf{70.2K} & \textbf{68.3K} & +1.9K\\ \hline
		Rnd. 19 (5 c)	& DE  & ES & YES & \textbf{83.1K} & 93.9K&  -10.8K & 83.9K  & \textbf{91.2K} & -7.3K  & 99.1K & 95.9K & +3.2K \\ \hline
		Rnd. 20 (5 c)	&  PSO & PSO & NO & 84.1K & 90.9K& -6.8K   &  85.9K & 93.2K &  -7.3K & \textbf{83.5K} & \textbf{87.3K} & -3.8K\\ \hline \hline
	\end{tabular}}
	\setlength{\tabcolsep}{5em}
	
	\label{table:randomonly}
	
\end{table}

\begin{table}
	\vspace{-3cm}	
	\caption {Predicted and actual most suited algorithm type/required FEN for Sphere function with linear constraint(s). The prediction model is trained with combination of pareto front and random points (PFR-PM) from multi objective evolver population. DE hard/easy (1 C) is a problem instance that is hard/easy for DE algorithm but easy/hard for the others.}
	{%
\hspace{-4cm}
	\begin{tabular}{|p{2.1cm}|p{1.2cm}|p{0.8cm}|p{0.6cm}|p{1.2cm}|p{1cm}|p{1.0cm}|p{1.2cm}|p{0.9cm}|p{1.0cm}|p{1.2cm}|p{1.2cm}|p{1.0cm}|}
	
	\hline \hline
	Instances name	& Predicted alg. & Actual alg. & Error & Predicted FEN for DE & Actual FEN for DE & Error for DE &Predicted FEN for ES & Actual FEN for ES& Error for ES & Predicted FEN for PSO & Actual FEN for PSO & Error for PSO  \\ \hline

	DE hard (1 c)&  ES & ES & NO & 83.8K & 86.3K & -2.5K&  \textbf{40.2K} & \textbf{41.5K} &  -1.3K& 45.8K & 43.2K & 2.6K  \\ \hline
	ES hard (1 c)& DE  & DE & NO & \textbf{43.6K} & \textbf{45.7K} & -2.1K  & 80.2K & 84.2K &  -4.0K & 45.1K & 48.3K&-3.2K \\ \hline
	PSO hard (1 c)& DE  & DE & NO  & \textbf{39.1K} & \textbf{37.2K} & 1.9K &42.4K & 41.8K & 0.6K  &78.1K & 80.1K &-2.0K  \\ \hline
	DE hard (2 c)&  PSO & ES & YES & 85.2K & 87.4K & -2.2K & 43.2K & \textbf{45.2K} & -2.0K& \textbf{41.9K} & 43.6K & -1.7K\\ \hline
	ES hard (2 c)&  DE & DE & NO & \textbf{45.8K} & \textbf{46.4K} & -0.6K  & 85.3K & 88.3K & -3.0K &48.9K & 46.8K& 2.1K\\ \hline
	PSO hard (2 c)& DE  & DE & NO & \textbf{41.8K} & \textbf{43.6K} & -1.8K &  47.9K & 45.1K &2.8K  & 85.3K & 83.2K & 2.1K\\ \hline
	DE hard (3 c)& ES  & ES & NO & 89.5K & 92.4K & -2.9K  & \textbf{43.1K}  & \textbf{45.2K} & -2.1K & 44.2K & 47.2K & -3.0K\\ \hline
	ES hard (3 c)&  PSO & PSO & NO  & 47.8K &  51.2K  & -3.4K & 94.7K & 91.2K& 3.5K & \textbf{42.9K}  & \textbf{48.2K} & -5.3K\\ \hline
	PSO hard (3 c)& DE  & DE & NO & \textbf{46.8K} & \textbf{49.6K} & -2.8K & 50.2K  & 53.5K & -3.3K  & 92.5K & 89.5K & 3.0K \\ \hline
	DE hard (4 c)& ES  & ES  & NO  & 96.2K & 93.7K & 2.5K & \textbf{49.2K} & \textbf{47.2K} & 2.0K & 51.2K  & 50.2K &1.0K \\ \hline
	ES hard (4 c)&  PSO & PSO & NO  & 50.4K & 49.2K & 1.2K & 84.7K & 89.1K &-4.4K & \textbf{46.8K} & \textbf{48.9K} & -2.1K\\ \hline
	PSO hard (4 c)&  ES & ES  & NO &\textbf{50.3K}  & 52.7K & -2.4K & 53.6K & \textbf{50.8K} &2.8K  & 90.2K & 87.5K &2.7K \\ \hline
	DE hard (5 c)& PSO  & ES & YES & 97.2K  & 96.3K & 0.9K& 55.9K & \textbf{53.3K} & 2.6K & \textbf{52.4K}  & 55.3K  & -2.9K \\ \hline
	ES hard (5 c)& DE  & DE & NO & \textbf{52.5K} & \textbf{53.8K} & -1.3K & 94.2K & 92.6K& 1.6K &53.8K  & 54.7K & -0.9K	\\ \hline
	PSO hard (5 c)& DE  & DE & NO & \textbf{50.2K} & \textbf{51.3K} & -1.1K & 56.3K & 55.2K &1.1K & 90.5K & 93.2K & -2.7K\\ \hline \hline
	
	DE easy (1 c)&  DE  & DE & NO & \textbf{46.3K}  & \textbf{48.9K} & -2.6K & 82.7K & 85.2K& -2.5K & 76.2K &79.1K & -2.9K\\ \hline
	ES easy (1 c)& ES & ES & NO & 86.6K & 81.4K & 5.2K & \textbf{48.2K} & \textbf{54.7K} & -6.5K & 80.2K & 81.7K & -1.5K\\ \hline
	PSO easy (1 c)& PSO & PSO & NO  & 86.3K & 87.1K &-0.8K  & 82.4K  & 84.2K&-1.8K &\textbf{45.7K} &\textbf{41.9K} &3.8K\\ \hline
	DE easy (2 c)&  DE & DE & NO & \textbf{48.1K} & \textbf{44.5K} & 3.6K & 87.4K & 83.1K & 4.3K & 79.4K & 81.6K & -2.2K\\ \hline
	ES easy (2 c)&  ES & ES & NO & 77.2K  & 83.6K &-6.4K & \textbf{52.3K} & \textbf{55.3K}&-3.0K & 77.4K & 78.9K &-1.5K \\ \hline
	PSO easy (2 c)& PSO  & PSO & NO & 83.2K & 81.4K & 1.8K & 78.4K &83.9K & -5.5K & \textbf{48.1K} & \textbf{49.4K} & -1.3K  \\ \hline
	DE easy (3 c)&  DE & DE & NO & \textbf{42.5K} & \textbf{48.4K} &  -5.9K & 86.8K & 89.4K &-2.6K  &87.4K & 84.5K & 2.9K \\ \hline
	ES easy (3 c)&  PSO & ES & YES & 90.9K & 87.8K & 3.1K & 48.9K & \textbf{46.2K} & 2.7K & \textbf{46.9K} & 48.0K & -1.1K\\ \hline
	PSO easy (3 c)& PSO & PSO & NO & 87.8K  & 89.5K  & -1.7K & 92.4K & 87.3K  & 5.1K & 47.0K & \textbf{49.1K} & -2.1K\\ \hline
	DE easy (4 c)&  DE & DE & NO & \textbf{50.2K}  & \textbf{53.2K} & -3.0K & 83.4K  &85.1K & -1.7K & 91.8K & 93.8K& -2.0K \\ \hline
	ES easy (4 c)&  ES & ES & NO & 86.7K & 89.4K & -2.7K & \textbf{50.8K} & \textbf{48.9K}& 1.9K  & 89.8K & 87.2K& 2.6K\\ \hline
	PSO easy (4 c)& DE  & DE & NO &  56.7K  & \textbf{55.5K} & 1.2K& 87.9K & 92.4K& -4.5K  & \textbf{55.2K}  & 68.9K& -13.7K\\ \hline
	DE easy (5 c)&  DE & DE & NO  & \textbf{55.9K} & \textbf{57.8K} & -1.9K & 93.1K & 95.3K & -2.2K & 82.4K & 81.5K& 0.9K\\ \hline
	ES easy (5 c)&  ES & ES & NO & 89.4K & 91.9K & -2.5K & \textbf{56.3K} & \textbf{55.7K}&  0.6K& 85.9K & 81.1K & 4.8K\\ \hline
	PSO easy (5 c)& PSO  & PSO & NO & 92.4K & 91.3K & 1.1K & 84.1K & 82.9K& 1.2K & \textbf{56.1K} & \textbf{57.6K} & -1.5K\\ \hline \hline

	Rnd. 1 (1 c)& PSO  & PSO & NO & 63.4K  & 65.7K   & -2.3K  &  55.4K & 56.9K & -1.5K & \textbf{51.7K} & \textbf{53.2K} & -1.5K  \\ \hline
	Rnd. 2 (1 c)& DE & DE & NO &\textbf{64.6K} & \textbf{61.9K}   & +2.7K &  64.7K & 64.2K & +0.5K  & 65.7K & 65.2K & +0.5K  \\ \hline
	Rnd. 3 (2 c)&  DE & DE & NO & \textbf{55.0K} & \textbf{59.8K}   & -4.8K & 62.3K  & 65.3K & -3.0K & 67.1K & 69.6K & -2.5K  \\ \hline
	Rnd. 4 (2 c)& ES  & ES &  NO & 68.6K & 72.2K & -3.6K & \textbf{61.4K}  & \textbf{59.4K} & +2.0K & 68.2K & 71.4K & -3.2K   \\ \hline
	Rnd. 5 (3 c)& DE  & DE & NO & \textbf{61.9K} & \textbf{65.4K}   & -3.5K &  62.5K & 66.9K & -4.4K  & 70.8K & 74.6K & -3.8K  \\ \hline
	Rnd. 6 (3 c)& PSO  & PSO & NO & 78.3K & 82.4K & -4.1K & 73.9K  & 78.3K & -4.4K  & \textbf{57.8K} & \textbf{61.4K} & -3.6K  \\ \hline
	Rnd. 7 (4 c)&  ES & ES & NO &86.4K & 83.2K   & +3.2K &  \textbf{69.4K} & \textbf{65.8K} & +3.6K & 74.7K & 78.9K & -4.2K  \\ \hline
	Rnd. 8 (4 c)&  ES & DE & YES & 74.6K & \textbf{71.1K}   & +3.5K & \textbf{73.7K}  & 79.2K & -5.5K & 75.8K & 73.2K & +2.6K  \\ \hline
	Rnd. 9 (5 c)&  DE & DE &  NO &\textbf{84.2K} & \textbf{82.7K} & +1.5K &  87.6K & 93.8K & -6.2K  & 94.2K & 97.3K & -3.1K  \\ \hline
	Rnd. 10 (5 c)&  PSO & PSO & NO & 83.7K & 93.2K   & -9.5K &  84.3K & 89.5K & -5.2K & \textbf{65.8K} & \textbf{68.4K} & -2.6K  \\ \hline
	\hline
	Rnd. 11 (1 c)& DE & DE & NO & \textbf{48.5K} & \textbf{44.6K} & +3.9K &  55.8K & 59.2K & -0.4K & 60.9K & 64.2K & -3.3K  \\ \hline
	Rnd. 12 (1 c)&  DE & DE & NO & \textbf{48.7K} & \textbf{45.2K} & +3.5K & 53.6K  & 58.9K & -5.3K & 63.6K & 69.4K & -5.8K  \\ \hline
	Rnd. 13 (2 c)& PSO  & PSO & NO & 55.7K & 61.2K & -5.5K &  65.9K & 64.3K & +1.6K & \textbf{54.9K} & \textbf{59.1K} & -4.2K  \\ \hline
	Rnd. 14 (2 c)& PSO  & PSO & NO & 76.0K & 79.1K & -3.1K & 66.6K  & 75.3K & -8.7K & \textbf{61.7K} & \textbf{63.4K} &  -1.7K \\ \hline
	Rnd. 15 (3 c)&  ES & PSO & YES &  68.8K& 71.9K & -3.1K  &  \textbf{62.7K} & 79.3K & -16.6K & 63.1K & \textbf{65.3K} & -2.2K  \\ \hline
	Rnd. 16 (3 c)& DE  & DE & NO & \textbf{66.5K} & \textbf{69.2K} & -2.7K &  72.2K & 79.3K & -7.1K & 71.5K & \textbf{70.1K} & +3.4K  \\ \hline
	Rnd. 17 (4 c)& ES & ES & NO & 70.4K & 73.9K & -3.5K & \textbf{70.3K}  & \textbf{75.9K} & -5.6K & 71.6K & 77.4K &  -5.8K \\ \hline
	Rnd. 18 (4 c)& PSO& PSO & NO & 75.4K & 85.3K & -9.9K &  83.5K & 78.3K & +5.2K & \textbf{71.2K} & \textbf{68.3K} & +2.9K  \\ \hline
	Rnd. 19 (5 c)& DE & ES & YES & \textbf{86.9K} & 93.9K & -7.0K &  88.5K & \textbf{91.2K} & -2.7L & 97.4K & 95.9K & -1.5K  \\ \hline
	Rnd. 20 (5 c)&  PSO & PSO & NO &  87.4K & 90.9K & -3.5K  & 90.3K  & 93.2K & -2.9K & \textbf{85.9K} & \textbf{87.3K} &  -1.4K \\ \hline
	
\end{tabular}}
\setlength{\tabcolsep}{5em}

\label{table:paretrandom}

\end{table}
	
\begin{table}	  	
	\caption{Comparison of PFR-PM with RO-PM models for Sphere function with various types of constraints (linear, quadratic and their combination). Average deviation of FEN denotes the average of differences between actual and predicted required FEN for PFR-PM and RO-PM.}
	{%
\hspace{-2cm}  	
\begin{tabular}{|p{4cm}|p{2cm}|p{2cm}|p{2cm}|p{2cm}|p{1cm}|}
	\hline \hline
	Problem & Success Rate RO-PM  & Success Rate FR-PM & Average deviation of FEN for RO-PM &  Average deviation of  FEN for PFR-PM & P value\\ \hline
	Sphere, 1lin & 2 & 26 &7.8K &2.4K & 0.004\\ \hline
	Sphere, 2lin & 1 & 27 & 8.6K & 1.8K & 0.013\\ \hline
	Sphere, 3lin & 1 & 26 & 12.6K & 3.2K & 0.018\\ \hline
	Sphere, 4lin & 5 & 28 & 13.6K & 3.5K & 0.006\\ \hline
	Sphere, 5lin & 1 & 28 & 17.4K & 2.7K & 0.028\\ \hline \hline
	Sphere, 1Quad & 1  & 27  & 11.9K & 2.1K & 0.035\\ \hline
	Sphere, 2Quad & 1 & 26 & 13.4K &  2.6K & 0.038\\ \hline
	Sphere, 3Quad & 3 & 28 & 15.8K & 3.7K & 0.043\\ \hline
	Sphere, 4Quad & 1 & 29 & 19.3K & 3.1K & 0.026\\ \hline
	Sphere, 5Quad & 5 & 28 & 21.6K & 4.3K & 0.035\\ \hline \hline
	Sphere, 4Lin, 1Quad & 2  & 24  & 13.4K  & 2.5K & 0.007\\ \hline
	Sphere, 3Lin, 2Quad & 2  & 26  & 13.8K & 2.8K & 0.004\\ \hline
	Sphere, 2Lin, 3Quad & 3  & 28  & 16.1K & 3.8K & 0.031\\ \hline
	Sphere, 1Lin, 4Quad & 5  & 27  & 18.9K & 3.7K & 0.016\\ \hline \hline
	
\end{tabular}}
\setlength{\tabcolsep}{5em}

\label{table:benchmarksphere}

\end{table}

\begin{table}	  	
	\caption{Comparison of PFR-PM with RO-PM models for Ackley function with various types of constraints (linear, quadratic and their combination). Average deviation of FEN denotes the average of differences between actual and predicted required FEN for PFR-PM and RO-PM.}
	{%
		\hspace{-2cm}  	
		\begin{tabular}{|p{4cm}|p{2cm}|p{2cm}|p{2cm}|p{2cm}|p{1cm}|}
			\hline \hline
			Problem & Success Rate RO-PM  & Success Rate FR-PM & Average deviation of FEN for RO-PM &  Average deviation of  FEN for PFR-PM & P value\\ \hline
			Ackley, 1lin & 0 & 27 & 9.3K & 2.5K & 0.043\\ \hline
			Ackley, 2lin & 1 & 27 & 11.5K  & 3.2K  & 0.016\\ \hline
			Ackley, 3lin & 2 & 25 & 10.3K  & 2.7K  & 0.004\\ \hline
			Ackley, 4lin & 1 & 28 & 14.7K  & 3.6K  & 0.008\\ \hline
			Ackley, 5lin & 6 & 29 & 13.8K  & 4.5K  & 0.025\\ \hline \hline
			Ackley, 1Quad & 2  & 29 & 16.3K  & 3.5K  & 0.046\\ \hline
			Ackley, 2Quad & 1 & 27 & 17.7K  & 4.1K   & 0.026\\ \hline
			Ackley, 3Quad & 0 & 25 & 18.3K  & 3.7K  & 0.043\\ \hline
			Ackley, 4Quad & 3 & 27 & 16.9K  & 5.1K  & 0.048\\ \hline
			Ackley, 5Quad & 2 & 29 & 21.9K  & 5.8K  & 0.034\\ \hline \hline
			Ackley, 4Lin, 1Quad & 4  & 24  &  15.8K  & 4.2K & 0.032 \\ \hline
			Ackley, 3Lin, 2Quad & 2  & 26  & 16.7K  & 4.6K  & 0.012 \\ \hline
			Ackley, 2Lin, 3Quad & 3  & 24  & 16.8K  & 4.9K  & 0.006 \\ \hline
			Ackley, 1Lin, 4Quad & 0  & 28  &19.8K  & 4.3K  & 0.021  \\ \hline \hline
			
		\end{tabular}}
		\setlength{\tabcolsep}{5em}
		
		\label{table:benchmarkackley}
		
	\end{table}
						
\begin{table}	
	
	\caption{Comparison of PFR-PM with RO-PM models for Rosenbrock function with various types of constraints (linear, quadratic and their combination). Average deviation of FEN denotes the average of differences between actual and predicted required FEN for PFR-PM and RO-PM.}
	{%
		\hspace{-2cm}  	
		\begin{tabular}{|p{4cm}|p{2cm}|p{2cm}|p{2cm}|p{2cm}|p{1cm}|}
			\hline \hline
			Problem & Success Rate RO-PM  & Success Rate FR-PM & Average deviation of FEN for RO-PM &  Average deviation of  FEN for PFR-PM & P value\\ \hline
			Rosenbrock, 1lin & 2 & 26 & 10.3K & 3.3K & 0.038 \\ \hline
			Rosenbrock, 2lin & 0 & 26 & 11.5K & 4.6K  & 0.035\\ \hline
			Rosenbrock, 3lin & 3 & 25 & 12.7K  & 3.6K & 0.002\\ \hline
			Rosenbrock, 4lin & 4 & 27 & 15.8K  & 5.2K  & 0.035\\ \hline
			Rosenbrock, 5lin & 5 & 28 &  19.4K & 5.1K  & 0.028\\ \hline \hline
			Rosenbrock, 1Quad & 2  & 27 & 17.4K  & 4.1K  & 0.017\\ \hline
			Rosenbrock, 2Quad & 1 & 29 &  21.5K & 4.7K   & 0.043\\ \hline
			Rosenbrock, 3Quad & 4 & 26 & 21.3K  & 5.7K  & 0.037\\ \hline
			Rosenbrock, 4Quad & 3 & 28 & 18.5K  & 5.2K  & 0.043\\ \hline
			Rosenbrock, 5Quad & 2 & 28 & 24.6K  & 6.9K  & 0.004\\ \hline \hline
			Rosenbrock, 4Lin, 1Quad & 1  & 28   &  14.7K  & 3.6K & 0.004 \\ \hline
			Rosenbrock, 3Lin, 2Quad & 0  & 24  &  17.4K & 4.7K  & 0.024 \\ \hline
			Rosenbrock, 2Lin, 3Quad & 4  & 25  & 19.5K  & 3.6K  & 0.029 \\ \hline
			Rosenbrock, 1Lin, 4Quad & 2  & 27  & 21.3K & 5.2K  &  0.006\\ \hline \hline
			
		\end{tabular}}
		\setlength{\tabcolsep}{5em}
		
		\label{table:benchmarkrosenbrock}
		
	\end{table}

	\end{document}